\documentclass[runningheads]{llncs}

\usepackage{eccv}

\usepackage{eccvabbrv}

\usepackage{graphicx}
\usepackage{booktabs}

\usepackage[accsupp]{axessibility}  
\makeatletter
\def\thanks#1{\protected@xdef\@thanks{\@thanks
        \protect\footnotetext{#1}}}
\makeatother

\usepackage{hyperref}
\usepackage{orcidlink}
\usepackage{multirow,multicol}
\usepackage{tabularx}
\usepackage{marvosym}
\begin{document}

\title{Stepping Stones: A Progressive Training Strategy for Audio-Visual Semantic Segmentation} 

\titlerunning{Stepping Stones}

\author{Juncheng Ma\inst{1}\orcidlink{0009-0001-9027-3111}\and
Peiwen Sun\inst{2}\orcidlink{0009-0005-3016-8554}\and
Yaoting Wang\inst{3}\orcidlink{0009-0004-5724-5698}\and
Di Hu\textsuperscript{\Letter}\inst{3,4}\orcidlink{0000-0002-7118-6733}
\thanks{\textsuperscript{\Letter}Corresponding author.}}

\authorrunning{J.~Ma et al.}

\institute{University of Chinese Academy of Sciences\\
\email{majuncheng21@mails.ucas.ac.cn}\\
\and
Beijing University of Posts and Telecommunications\\
\email{sunpeiwen@bupt.edu.cn}\\
\and
Gaoling School of Artificial Intelligence, Renmin University of China, China\\
\email{yaoting.wang@outlook.com}\\ \email{dihu@ruc.edu.cn}\\
\and
Engineering Research Center of Next-Generation Search and Recommendation
}

\maketitle

\begin{abstract}
Audio-Visual Segmentation (AVS) aims to achieve pixel-level localization of sound sources in videos, while Audio-Visual Semantic Segmentation (AVSS), as an extension of AVS, further pursues semantic understanding of audio-visual scenes. However, since the AVSS task requires the establishment of audio-visual correspondence and semantic understanding simultaneously, we observe that previous methods have struggled to handle this mashup of objectives in end-to-end training, resulting in insufficient learning and sub-optimization. Therefore, we propose a two-stage training strategy called \textit{Stepping Stones}, which decomposes the AVSS task into two simple subtasks from localization to semantic understanding, which are fully optimized in each stage to achieve step-by-step global optimization. This training strategy has also proved its generalization and effectiveness on existing methods. To further improve the performance of AVS tasks, we propose a novel framework Adaptive Audio Visual Segmentation, in which we incorporate an adaptive audio query generator and integrate masked attention into the transformer decoder, facilitating the adaptive fusion of visual and audio features.  Extensive experiments demonstrate that our methods achieve state-of-the-art results on all three AVS benchmarks. The project homepage can be accessed at \href{https://gewu-lab.github.io/stepping_stones/}{https://gewu-lab.github.io/stepping\_stones}.

\keywords{Audio-Visual Segmentation \and Audio-Visual Semantic Segmentation \and Multimodality}
\end{abstract}

\section{Introduction}
\label{sec:intro}

In the real world, audio is consistently associated with its sources, enabling humans to locate sound sources based on what they hear. In the past few years, this phenomenon has spurred significant research on Audio-Visual Localization (AVL) \cite{local1,local2,wei2022learning} aiming to predict the location of sound sources within a video, enhancing the human-like perceptual capabilities of machines. However, due to the lack of precise labels, AVL methods tend to capture only a coarse contour of the sound source. Therefore, Audio-Visual Segmentation (AVS) \cite{avsbench,wang2024refavs,wang2024segpref} has recently been proposed as a more fine-grained task over AVL, providing pixel-level labels of sound sources. Subsequently, a more challenging task Audio-Visual Semantic Segmentation (AVSS)\cite{avss} is proposed in the pursuit of machine understanding of audio-visual scenes at a semantic level, requiring predicting semantic labels for sound sources. In contrast to the traditional Semantic Segmentation (SS) task, AVSS requires achieving audio-visual alignment utilizing the newly introduced audio modality. Compared to AVS, AVSS demands a deeper semantic understanding of the audio-visual scene. The heightened complexity of the objective in AVSS exacerbates the challenge for the model.
\begin{figure}[tb]
  \centering
  \includegraphics[width=0.9\linewidth]{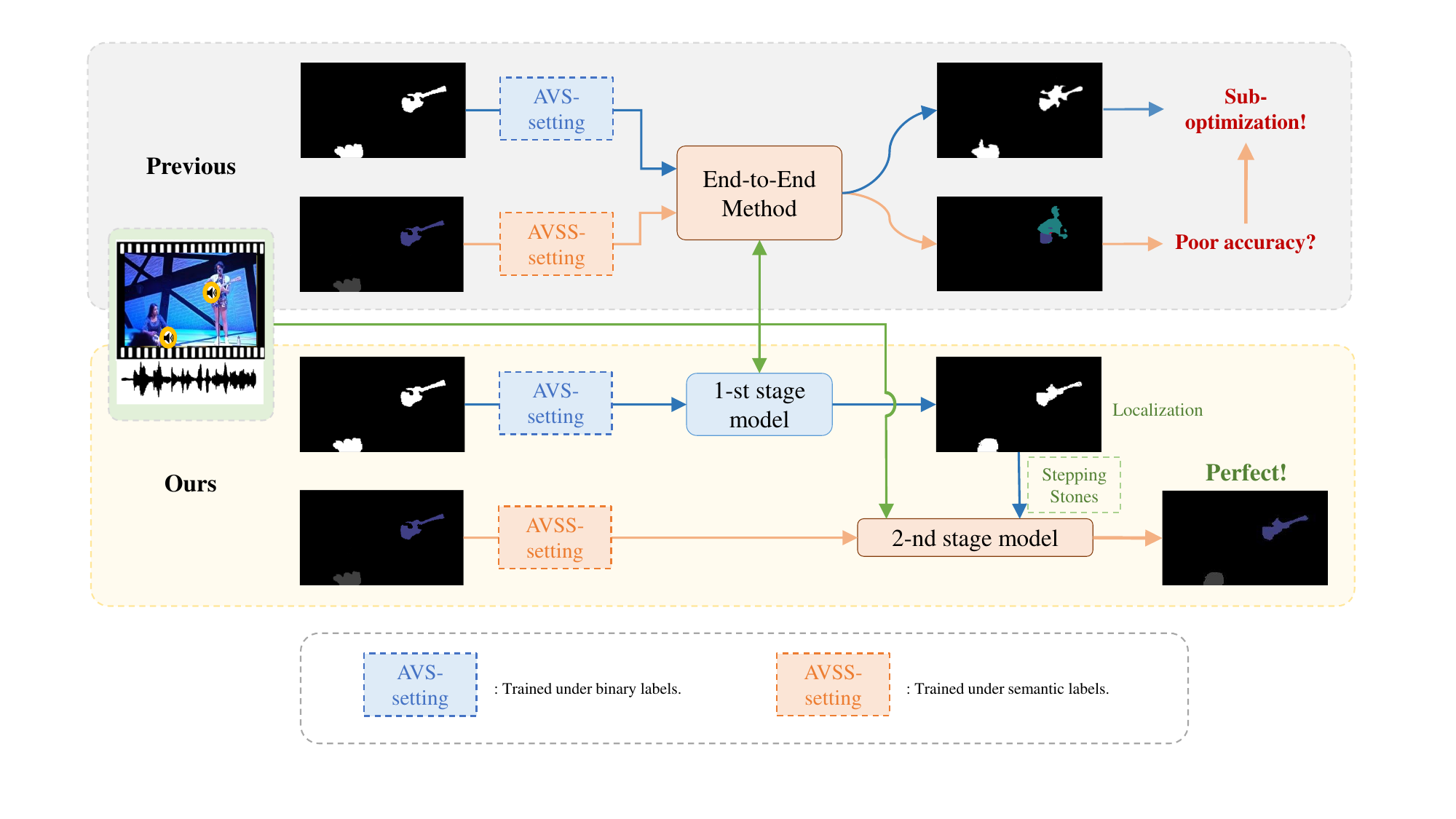}
  \caption{Comparison between previous methods and our \textit{Stepping Stones} training strategy. \textit{Top}: Previous end-to-end AVSS approaches result in sub-optimization on audio-visual alignment. Specifically, when trained under the AVSS-setting, these methods exhibit weaker sound source localization capability compared to those trained under the AVS-setting. \textit{Bottom:} our \textit{Stepping Stones} training strategy decomposes the intricate AVSS task into two relatively simple subtasks to be fully learned in two stages, enhancing the performance significantly. }
  \label{fig:teaser}
\end{figure}

AVSBench\cite{avsbench} is the first framework for AVS that employs multimodal feature fusion to achieve dense predictions. Subsequent advancements include the integration of audio-queried transformer structures to alleviate receptive field constraints of convolutional approaches \cite{aqformer, avsegformer}, and the incorporation of temporal contexts to capture audio-visual spatial-temporal correlations\cite{catr}. In addition, \cite{SAMA,gavs} use foundation models such as SAM\cite{sam} to improve the generalization and performance of the model. 
However, previous methods may have resulted in sub-optimal outcomes and inadequate learning, attributed to the inherent complexity of the AVSS task and the indiscriminate propagation of supervised signals across different modalities during end-to-end training.

Specifically, the AVSS model necessitates semantic-level scene comprehension while simultaneously achieving audio-visual alignment to predict semantic labels of sound sources, effectively combining aspects of AVS and SS. Although the end-to-end approach is widely favored for its convenience, previous end-to-end methods for the AVSS task, while aiming for global optimization, may not sufficiently learn or reach sub-optimization under the supervision of such mixed objectives due to the following two aspects. Firstly, supervisory signals are ambiguous during end-to-end optimization. For example, a pixel labeled as \textit{background} may either truly belong to the \textit{background} or represent a silent but potentially semantic object such as a \textit{guitar}. Yet, distinguishing between these scenarios during training is not feasible, potentially leading to confusion in semantic understanding when the \textit{background} signal propagates to the entire model in the latter case. On the other hand, the propagation of supervisory signals during end-to-end training may be insufficiently utilized. In previous end-to-end approaches, the model needs to simultaneously learn semantic understanding primarily relying on visual cues and audio-guided modal alignment to filter out silence regions, each having different focal points, potentially resulting in sub-optimization respectively while striving for global optimization. Limitation of end-to-end methods was also corroborated through experiments conducted on AVSBench \cite{avsbench}, as depicted in the \textit{top} of \cref{fig:teaser}, wherein we observed that the model trained under the AVSS task setting exhibits inferior audio-visual alignment (with semantics disregarded) compared to the model trained under the AVS task setting. More results are shown in \cref{fig:compar}. 

To solve this problem and boost the performance of previous AVSS methods, we propose a simple yet effective two-stage progressive training strategy called \textit{Stepping Stones} (depicted in the \textit{bottom} of \cref{fig:teaser}) to decompose the intricate AVSS task into two relatively simple subtasks to be fully learned in the two stages. In the first stage, the model is trained using binary labels without semantic information, prioritizing the alignment of audio and visual modalities to establish fine-grained correlations at the pixel level.  Subsequently, in the second stage, the model is trained with semantic labels as supervision, leveraging the sound source localization results from the first stage as stepping stones. These stepping stones serve as a shortcut for the model to readily recognize sounding pixels during training, thus focusing on the semantic understanding of the sound source. This approach is generalizable and can be readily applied to existing end-to-end AVSS models. Moreover, acknowledging the considerable error between the first stage results and ground truth during inference, we also introduce the Robust Audio-aware Key Generator to robustly integrate auditory information into visual features as keys in the cross-modal cross-attention module. Since the second stage leverages the sound source localization results as prior knowledge, the model can focus solely on the semantic comprehension of a given sound source region, thus overcoming the challenge of ambiguous supervision signals during the end-to-end learning. Furthermore, we comprehensively train audio-visual alignment and visually dominant semantic understanding in two separate training phases, which facilitates the optimization of both subtasks of the model, leading to improved global optimization.

In addition, we propose a novel audio-queried transformer model built upon Mask2Former\cite{mask2former}, termed Adaptive Audio Visual Segmentation (AAVS). Our framework comprises two key innovations compared with previous methods\cite{avsegformer,aqformer,catr,autr}. Firstly, in previous works, audio queries were either nondiscriminatory or were generated using a module with intricate parameters. We propose an adaptive audio query generator to encode audio queries adaptively, without introducing additional parameters. This approach enhances queries to concentrate on pertinent audio features, thereby improving model performance. Secondly, inspired by \cite{mask2former}, we integrate masked attention into the transformer decoder, enabling the model to dynamically adjust its focus on visual features during decoding, which promotes faster convergence and enhanced performance.

To summarise, our main contributions are as follows.
\begin{itemize}
    \item[$\bullet$] We propose a simple yet effective progressive
    training strategy, \textit{Stepping Stones}, for audio-visual semantic segmentation, which can be readily applied to all existing methods. Experiments on several methods demonstrate a notable improvement in validating the strategy's generalization.
    \item[$\bullet$] We propose Adaptive Audio Visual Segmentation (AAVS), a novel framework for audio-visual segmentation. AAVS incorporates an adaptive audio query generator and integrates masked attention into the transformer decoder, facilitating the adaptive fusion of visual and audio features.
    \item[$\bullet$] Extensive experiments show that AAVS outperforms existing approaches on the AVSBench-object dataset. Moreover, AAVS trained using our proposed \textit{Stepping Stones} strategy, exhibits considerable performance gains over previous methods on the more challenging AVSBench-semantic dataset.
\end{itemize}

\section{Related Work}
\subsection{Audio-Visual Localization}
Audio-Visual Localization (AVL) aims to predict the location of sound sources within a video, with results typically presented as coarse heatmaps\cite{local1,local2}. Prior investigations\cite{local3,local4} have employed Audio-Visual Correspondence as a self-supervised learning objective, leveraging the inherent alignment between audio and visual features along the temporal dimension. Further advancements by \cite{local5,local6} have introduced contrastive learning, utilizing positive and negative instances to enhance sound source localization, emerging as a dominant approach in recent years. To tackle the complexities of multi-source localization, \cite{local7} proposed a two-stage training framework incorporating objectives ranging from image-level to class-level granularity. The method facilitates the progressive refinement of multi-source localization, serving as a foundational approach in this field. These pioneering studies have significantly influenced our proposal of a two-stage framework, achieving a step-by-step optimization from localization to semantic understanding for Audio-Visual Semantic Segmentation.

\subsection{Audio-Visual Segmentation}
In contrast to AVL, Audio-Visual Segmentation (AVS) represents a more fine-grained task introduced by \cite{avsbench}. AVS offers pixel-level annotations for sound source localization, aiming to achieve a detailed understanding of audio-visual scenes, encompassing both single-source and multi-source scenarios. Furthermore, Audio-Visual Semantic Segmentation emerges as a more challenging task\cite{avss}, recently proposed as an extension of AVS, with the goal of attaining fine-grained sound source localization and semantic comprehension simultaneously. Existing methods primarily rely on the fusion of visual and audio features. AVSBench\cite{avsbench} is the first framework that employs multimodal feature fusion to achieve pixel-level dense predictions. Subsequent advancements \cite{aqformer, avsegformer} have introduced an audio-queried transformer structure to establish global contextual dependencies, thereby mitigating limitations of the receptive field inherent in convolutional approaches. Other methods include the incorporation of temporal contexts to capture audio-visual spatial-temporal correlations\cite{catr}, the utilization of bi-directional generation to establish robust correlations \cite{avsbg}, and the integration of generative models to facilitate audio-visual fusion\cite{diffuse}. Additionally, several works\cite{SAMA,gavs} leverage the power of large foundation models such as SAM\cite{sam} to implement audio-visual segmentation and enhance model generalization. However, previous approaches have overlooked the intricacies inherent in the AVSS task, employing end-to-end models that face challenges in comprehensively mastering both audio-visual alignment and semantic comprehension. Consequently, we propose a two-stage strategy, aiming to decompose the AVSS task into two simple subtasks from localization to semantic understanding, thereby facilitating incremental progress toward enhanced model performance.
\subsection{Semantic Segmentation}
Semantic segmentation is a fundamental task in computer vision, involving the assignment of semantic labels to each pixel in an image. Early methodologies \cite{fcn,unet} employ an encoder-decoder architecture and adopt per-pixel dense prediction, which entails downsampling to aggregate global semantic features during encoding and upsampling to recover local detailed features during decoding. With the advent of vision transformers\cite{dosovitskiy2021imageworth16x16words,swin}, recent works\cite{jain2022oneformertransformerruleuniversal,mask2former,li2024omgsegmodelgoodsegmentation} have strived for universal segmentation, including semantic segmentation, instance segmentation, and image segmentation, outperforming previous specialized models. Among them, Mask2Former\cite{mask2former} proposes per-mask classification, which predicts numerous masks and assigns semantic labels to each one. This approach not only unifies semantic segmentation with other image segmentation tasks but also achieves noteworthy performance. Motivated by these advancements, we propose a novel model using per-mask classification for audio-visual segmentation designed to adaptively encode audio queries and dynamically capture visual features using masked attention to attain state-of-the-art performance.
\section{Methods}
In this section, we will first introduce our proposed AAVS framework in \cref{sec:aavs}, including the Adaptive Audio Query Generator, which dynamically fuses audio features with object queries, and the transformer decoder with masked attention to flexibly adjust attention region of the visual feature maps for audio queries. In addition, we will illustrate our training strategy \textit{Stepping Stones} in \cref{sec:strategy} and show methods to apply it efficiently to the AAVS model in detail.

\subsection{Adaptive Audio Visual Segmentation}
\label{sec:aavs}
\subsubsection{Feature Extraction.} 
\label{sec:extraction}
Followed \cite{avsbench,catr}, we employ pre-trained backbone as visual encoder to extract feature maps across different scales. After that, we utilize the Feature Pyramid Network\cite{fpn} as a pixel decoder to aggregate feature maps of various resolutions. To integrate the global semantic and the fine-grained features, we iteratively aggregate neighboring resolution feature maps, resulting in merged multi-scale feature  $F_V^{i}$$\in \mathbb{R}^{T \times D \times \frac{H}{2^{i+1}} \times \frac{W}{2^{i+1}}}, i \in \{0,1,2,3\}$. Here, $T$, $D$, $H$ and $W$ denotes the number of frames, embedding dimensions (default to 256) and the original size respectively.

We preprocess the audio input following previous work\cite{avsbench,gavs}. Then, we utilize VGGish\cite{vggish} pre-trained on Audioset\cite{audioset} and a linear projector to extract audio features as $F_A\in \mathbb{R}^{T\times D}$.
\begin{figure}[tb] 
  \centering
  \includegraphics[width=1.0\linewidth]{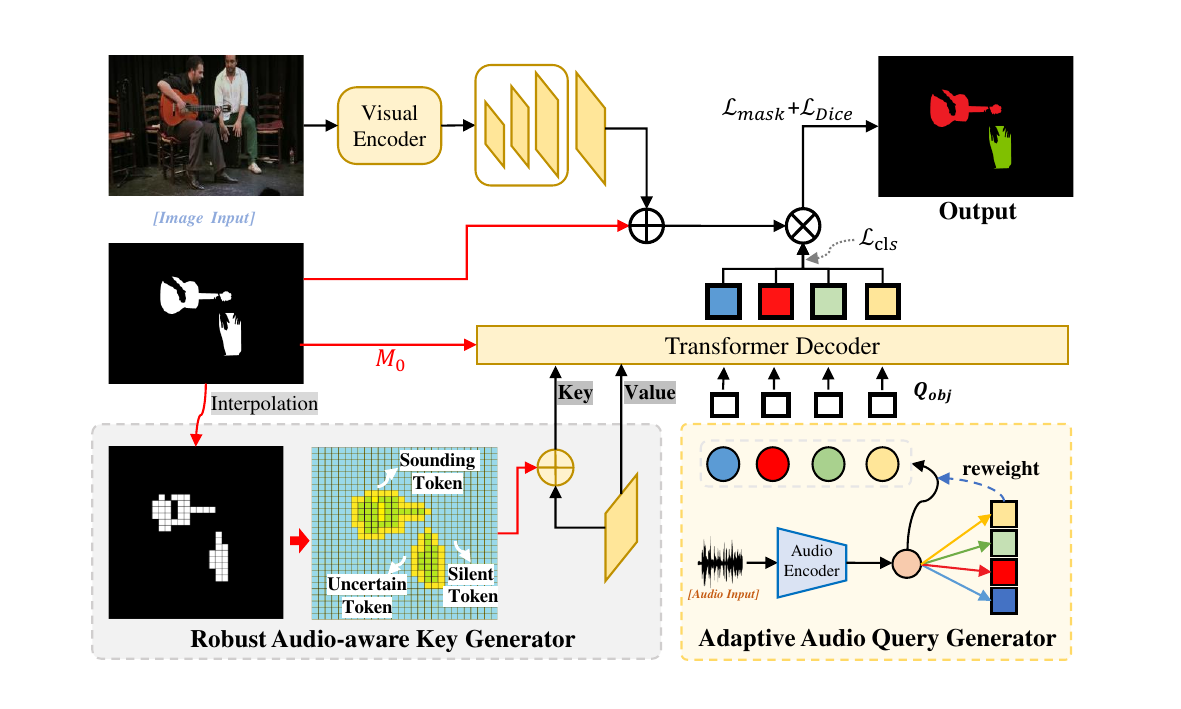}
  \caption{Overview of AAVS framework.  (1) Visual and audio features are extracted by the pre-trained encoder; (2) Adaptive Audio Query Generator is proposed to generate audio queries; (3) In the transformer decoder, audio-aware queries are integrated with visual feature maps, and masked cross-attention facilitates queries to dynamically adjust the attention range; (4) Finally, refined queries are merged with the mask feature to obtain the final prediction mask. \textcolor{red}{Red arrows} indicate newly introduced methods when implementing the \textit{Stepping Stones} strategy.}
  \label{fig:model}
\end{figure}
\subsubsection{Adaptive Audio Query Generator.}
\label{sec:audio}
Most of existing methods\cite{catr,aqformer,autr} for generating audio queries lack discriminative features in relation to the audio, simply repeating the audio features $F_A$ and combining it with the learnable object queries to be fed into a transformer decoder then. A recent work\cite{avsegformer} implicitly decomposes the audio features to obtain the audio queries using a transformer structure. However, due to the inherent complexity and ambiguity of audio, we believe that this method does not effectively decompose audio components and is parameter redundant.

Therefore, we propose an Adaptive Audio Query Generator without introducing extra parameters to encode adaptive audio queries, which can directly replace the previous \textit{repeat} method.
Initially, we define $N_q$ learnable object queries $Q_{obj}\in \mathbb{R}^{N_q\times D}$ and $N_q$ corresponding audio prototypes $P_{audio}\in \mathbb{R}^{N_q\times D}$ to represent $N_q$ potential sound source objects. Each $P_{audio}^i  ( i\in [1, N_q] )$ also serves as the positional embedding for $Q_{obj}^i$ in the transformer decoder enhancing their correlation during training, which also demonstrates that the module introduces no additional parameters. 

Subsequently, we dynamically weight $F_A$ based on the cosine similarity with each $P_{audio}^i$ and then integrate the results with $Q_{obj}^i$ to obtain audio-conditioned query $Q_{a}^i\in \mathbb{R}^{D}$,
\begin{equation}
Q_{a}^i=\frac{<P_{audio}^i, F_A>}{|P_{audio}^i||F_A|}F_A+Q_{obj}^i.
  \label{eq:adaptive}
\end{equation}

By this method, the original audio feature $F_A$ can be adaptively fused to the potential object query $Q_{obj}^i$ according to corresponding audio prototype $P_{audio}^i$, improving the discriminative capacity of the model towards audio signals.
\subsubsection{Transformer Decoder.}
\label{sec:mask}
In the transformer decoder, we integrate $Q_{a}$ and multi-scale visual features [$F_V^{2}, F_V^{3}, F_V^{4}$] sufficiently to obtain refined $Q_{fuse}$. Specifically, the decoder consists of four stages, each comprising three transformer layers corresponding to distinct scales of feature maps. Within each layer, we first compute the cross-attention between $Q_{a}$ and $F_V^{i}$ using the Masked Multihead Attention\cite{mask2former}, which enables the model to focus more on the region to be predicted and thus adaptively adjust the attention range.  Subsequently, we perform self-attention on $Q_{a}$ and feed it into the Feed Forward Network. 
\subsubsection{Loss.}

After decoding, we adopt per-mask classification\cite{maskformer} to generate the final segmentation predictions. Specifically, for each $Q_{fuse}^i$, the mask predictor combines mask feature $F_V^1$ to predict a binary mask, while the class predictor determines the category for this mask. Then, we use the Hungarian algorithm\cite{mask2former} based on the loss function described in Eq. \ref{eq:loss} to identify the optimal set of queries that minimizes Eq. \ref{eq:loss} and back-propagate the loss,

\begin{equation}
\mathcal{L}_{main}=\lambda_{cls}\mathcal{L}_{cls}+\lambda_{mask}\mathcal{L}_{mask}+\lambda_{dice}\mathcal{L}_{dice}.
  \label{eq:loss}
\end{equation}

The loss function comprises three components: $\mathcal{L}_{cls}$ represents the cross-entropy loss for mask classification, while $\mathcal{L}_{mask}$ and $\mathcal{L}_{dice}$ denote the cross-entropy loss and Dice loss\cite{dice} for mask prediction respectively. We introduce $\lambda_{cls}$, $\lambda_{mask}$, and $\lambda_{dice}$ to weight these losses. 

Additionally, we employ deep supervision to compute auxiliary losses for the output of each transformer decoder layer to improve performance. The total loss is presented in Eq. \ref{eq:allloss}, 
\begin{equation}
\mathcal{L}=\mathcal{L}_{main}+\lambda_{aux}\mathcal{L}_{aux},
  \label{eq:allloss}
\end{equation}
where $\lambda_{aux}$ is used to weight the auxiliary loss.
\subsubsection{Inference.}
For inference, $Q_{fuse}$ is fed to the mask predictor and class predictor to generate mask prediction $O_{mask}\in \mathbb{R}^{T\times N\times \frac H4\times \frac W4}$ and class prediction $O_{class}\in \mathbb{R}^{T\times N\times (C+1)}$, where C donates the total number of classes and $1$ donates a extra class \textit{null}. Then, $O_{mask}$ and $O_{class}$ are postprocessed to obtain $O_{pred}\in \mathbb{R}^{T\times (C+1)\times \frac H4\times \frac W4}$ followed \cite{mask2former}. Finally, We apply bilinear interpolation to upsample $O_{pred}$ to its original size and take the argmax operation along the category dimension to obtain the final prediction mask $P\in \mathbb{R}^{T \times H\times W}$.

\subsection{Stepping Stones Training Strategy}
\label{sec:strategy}
As mentioned above, the objective of AVSS can be viewed as the combination of AVS and SS, which necessitates the model to simultaneously learn audio-visual correspondence and semantic understanding. However, the integration of these objectives has led in insufficient learning and sub-optimal performance in previous end-to-end methods, prompting us to propose a two-stage training strategy, termed \textit{Stepping Stones}, to tackle the intricate goal as depicted in \cref{fig:teaser}. In the first stage, we train an AVS model supervised by binary labels to focus on sound source localization. Subsequently, in the second stage, we train an AVSS model supervised by semantic labels, leveraging the first-stage results $\mathbf{\hat{M}}$ as stepping stones. Here, the model shifts its emphasis to semantic discrimination of sound source, with audio augmenting visual modal features to enhance performance. By decomposing the AVSS task into two stages of step-by-step learning, we further maximize the capabilities of the end-to-end model and facilitate its comprehensive training. To mitigate the impact of errors, we used ground truth labels $\mathbf{M}$ during training and $\mathbf{\hat{M}}$ exclusively during testing.

Next, we will elaborate on how the second stage model effectively utilizes the first-stage results as stepping stones when applying the strategy to AAVS. The modified AVSS model is illustrated in \cref{fig:model}.

\subsubsection{Prior Knowledge Input as Stepping Stones.}
\label{par:prior}

In contrast to semantic segmentation, the newly introduced audio modality in the AVSS task plays a crucial role in filtering out silent objects through modal alignment. Consequently, $\mathbf{M}$ offers reliable prior knowledge and mitigates the need for audio-visual alignment within the second-stage model. In the second stage, the model can stand on the stepping stones and concentrate further on comprehending visual semantics.

Specifically, we integrate $\mathbf{M}$ with the mask feature $F_V^1$ in the prediction head. This simple operation serves as a shortcut for the model, aiding in the easy recognition of the sounding region and enhancing its focus on the semantic discrimination of the sound source during training.

Additionally, we utilize $\mathbf{M}$ as the initialization mask for masked attention in the transformer decoder. Previous initialization methods result in notably low accuracy of the original mask, where the initial attention mask is derived from the initial queries. An inaccurate initialization mask can hinder the effective confinement of attention to the foreground region in the cross-attention module, potentially leading to slow convergence or model degradation performance. Therefore, we naturally employ $\mathbf{M}$ as prior information to initialize the attention mask, enabling queries to concentrate the region relevant to the sound source.
\subsubsection{Robust Audio-aware Key Generator.}
\label{sec:robust}
To mitigate the inevitable error between $\mathbf{\hat{M}}$ obtained from the first stage and ground truth labels $\mathbf{M}$, we devised the Robust Audio-aware Key Generator to fuse audio information into feature maps [$F_V^2, F_V^3, F_V^4$] as key in cross-modal cross attention to bolster the model's robustness and maximize its utilization of the stepping stones. 

The module incorporates three learnable embeddings: $E_{silent}$, $E_{uncertain}$, $E_{sounding}$ and two thresholds $\tau_1$, $\tau_2$ ($\tau_1<\tau_2$). Initially, $\mathbf{M}$ is resized to match the feature map $F_V^i$. Subsequently, $\mathbf{M}_{resized}$ is thresholded by $\tau_1$ and $\tau_2$ to produce a index mask. Then, an audio-aware embedding mask $\mathbf{M}_{audio}\in \mathbb{R}^{T  \times \frac{H}{2^{i+1}} \times \frac{W}{2^{i+1}} \times D}$ is generated based on the index mask, which is then combined with the $F_V^i$ to derive audio-aware keys for cross attention. The entire process is illustrated in Eq. \ref{eq:robust},
\begin{equation}
M_{audio}^{t,i,j} = 
    \begin{cases}
        E_{silent} & \mbox{if } \mathbf{M}_{resized}^{t,i,j} < \tau_1 \\
        E_{uncertain}   & \mbox{if } \tau_1 \leq \mathbf{M}_{resized}^{t,i,j} \leq \tau_2\\
        E_{sounding}& \mbox{if } \mathbf{M}_{resized}^{t,i,j} > \tau_2
    \end{cases},
  \label{eq:robust}
\end{equation}
where $t$ represents the current frame, and $(i, j)$ denotes the coordinates of the current pixel. The module enables the queries $Q_a$ to dynamically adjust the fusion strategy with visual features based on audio-aware keys during decoding. For instance, in a guitar video scenario, while the original visual key may solely represent the corresponding region as a guitar feature, the audio-aware key can offer supplementary details regarding whether the guitar is \textit{silent}, \textit{uncertain}, or \textit{sounding}, thereby enhancing the model's robustness and capacity.

\section{Experiments}
\label{exper}
\subsection{Datasets}
\subsubsection{AVSBench-Object}\cite{avsbench} is an audio-visual dataset for audio-visual segmentation. The dataset consists of two subsets evaluating the S4 and MS3 subtasks respectively, providing pixel-level binary labels about sound sources in video. The S4 subset contains 4,932 videos, with 3,452 videos for training, 740 for validation, and 740 for testing. Conversely, the MS3 subset includes 424 videos, with 286 videos for training, 64 videos for validation, and 64 videos for testing.

\subsubsection{AVSBench-Semantic}\cite{avss} introduces the AVSS subtask as an extension of AVSBench-Object. In addition to providing semantic labels for S4 and MS3 subsets, it also expands the V2 subset. Overall, the AVSBench-Semantic dataset consists of 8,498 videos for training, 1,304 videos for validation, and 1,554 videos for testing. The target objects span 71 categories, encompassing humans, musical instruments, animals, and tools.

\subsubsection{Evaluation Metrics.} Followed \cite{avsbench,avsegformer}, we use the mean Intersection-over-Union (mIoU) and F-score as the evaluation metrics.
\subsubsection{Implementation Details.} 
We choose the released Mask2Former\cite{mask2former} model with Swin-B\cite{swin} visual backbone pre-trained on ADE20k\cite{ade} as weight initialization. To effectively leverage the knowledge within the pre-trained models, we freeze most of the parameters and introduce adapters\cite{adapter} for training to achieve faster convergence and improved performance. We utilize the AdamW optimizer with a learning rate of $10^{-4}$. The batch size is set to 1. 
To evaluate the effectiveness and generalization of the \textit{Stepping Stones} strategy, we train and test AVSBench and AVSegformer model according to the setting of \cite{avsbench,avsegformer}.
\begin{table}[tb]
  \caption{Quantitative (mIoU, F-score) results on AVSBench dataset with transformer-based visual backbone. $^*$ indicates that the model uses the \textit{Stepping Stones} strategy.}
  \label{tab:result}
  \centering
  \begin{tabularx}{1.0\textwidth}{@{}l *{6}{X}l@{}}
        \toprule[0.8pt]
        \multirow{2}{*}{\textbf{Method}}  & \multicolumn{2}{c}{\textbf{S4}}  & \multicolumn{2}{c}{\textbf{MS3}}& \multicolumn{2}{c}{\textbf{AVSS}}&\multirow{2}{*}{\textbf{Reference}}\\
        \cmidrule(r){2-3}\cmidrule(r){4-5}\cmidrule(r){6-7}
        & mIoU & F-score & mIoU & F-score& mIoU & F-score&\\
        \midrule
        AVSBench\cite{avsbench}&78.7&87.9&54.0& 64.5&29.8&35.2 &ECCV'2022\\
        AVSC\cite{avsc} &80.6 &88.2&58.2&65.1&-&-&ACM MM'2023\\ CATR\cite{catr}&81.4&89.6&59.0&70.0&32.8&38.5&ACM MM'2023\\
        DiffusionAVS\cite{diffuse}&81.4&90.2&58.2&70.9&-&-&ArXiv'2023\\
        ECMVAE\cite{vae}& 81.7&90.1&57.8&70.8&-&-&CVPR'2023\\

        AuTR\cite{autr}&80.4&89.1 &56.2& 67.2& -&-&ArXiv'2023\\
        SAMA-AVS\cite{SAMA}&81.5&88.6&63.1&69.1&-&-&WACV'2023\\
        AQFormer\cite{aqformer}& 81.6&89.4&61.1&72.1&-&-&IJCAI'2023\\
        AVSegFormer\cite{avsegformer}&82.1&89.9&58.4&69.3&36.7&42.0&AAAI'2024\\
        AVSBG\cite{avsbg}&81.7&90.4&55.1&66.8&-&-&AAAI'2024\\
        GAVS\cite{gavs}&80.1&90.2&63.7&77.4&-&-&AAAI'2024\\
        MUTR\cite{mutr}&81.5 &89.8&65.0 &73.0&-&-&AAAI'2024\\

        \midrule
        AAVS (Ours)&\textbf{83.2} & \textbf{91.3}&\textbf{67.3}&\textbf{77.6}&\textbf{48.5$^*$}&\textbf{53.2$^*$}&ECCV'2024\\
        \bottomrule
    \end{tabularx}
\end{table}
\subsection{Results and Analysis}
We conduct experiments on all three subtasks (S4, MS3, AVSS) of the AVSBench dataset to evaluate the effectiveness of our method and compare it with previous approaches. For fairness, we
employ transformer-based backbone to extract visual features and the Audioset\cite{audioset} pre-trained VGGish\cite{vggish} to extract audio features for all methods. 

\subsubsection{Quantitative Comparison.} 
We conduct a comprehensive comparison between our AAVS model and existing methods on the AVSBench-Object dataset. The results are presented in \cref{tab:result}. Our AAVS model outperforms previous methods in terms of mIoU and F-score for both the S4 and MS3 subtasks. Specifically, for the S4 subtask, AAVS demonstrates performance improvements of 1.1\% in mIoU and 0.9\% in the F-score. For the MS3 subtask, AAVS showcases performance improvements of 2.3\% in mIoU and 0.2\% in the F-score. We attribute these enhancements to the efficacy of our proposed method, which enhances audio-visual modal alignment through adaptive fusion of visual and audio features. It is noteworthy that the improvement on the MS3 subtask is significantly higher than that on the S4 subtask, which we attribute to the greater complexity of the MS3 task scenarios and the adaptive nature of our approach in addressing multi-sound source localization challenges. 

For the AVSS subtask, we also conduct experiments to compare our approach with previous work. As depicted in \cref{tab:result}, when the \textit{Stepping Stones} training strategy is applied to the AAVS model, our model demonstrates performance enhancements of 11.8\% mIoU and 11.2\% F-score. This significant improvement arises partly from the capabilities of the AAVS model itself and partly from the \textit{Stepping Stones} training strategy, which aids the model in learning audio-visual alignment and semantic understanding more effectively.
\subsubsection{Qualitative Comparison}
\begin{figure}[tb]
  \centering
  \includegraphics[width=1\linewidth]{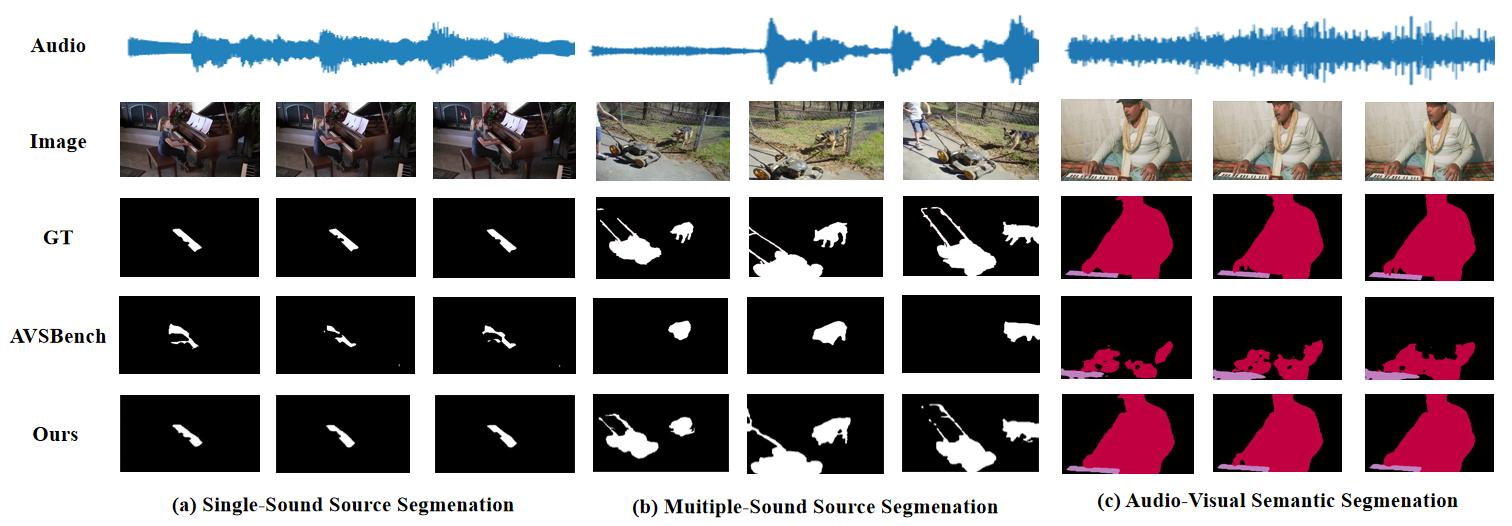}
  \caption{Qualitative comparison with previous methods. The left is from the S4 subtask, the center is from the MS3 subtask, and the right side is from the AVSS subtask. }
  \label{fig:qual}

\end{figure}

\cref{fig:qual} is a comparison of prediction results obtained from AVSBench and our proposed method. It is evident that our method excels in accurately localizing and segmenting sounding objects, aligning closely with ground truth labels. In the simple S4 task, AAVS effectively identifies the sound source and delineates its boundary with precision. In the more intricate MS3 task, AAVS locates multiple sources well from the audio and provides clear outlines for each. In the most challenging AVSS task, employing the \textit{Stepping Stones} training strategy enhances the performance of the AAVS model significantly, yielding segmentation results that closely approximate the ground truth. More qualitative comparison results can be seen in Supplementary Material.

\begin{table}[tb]
  \caption{Experiments on previous AVSS Methods. SS means \textit{Stepping Stones} strategy in this table. \textit{low}, \textit{high}, and \textit{oracle} corresponded to three levels of mIoU values of the first stage results. Among them, \textit{oracle} indicates ground truth labels.}
  \label{tab:general}
  \centering
  \begin{tabular}{@{}lllllllll@{}}
    \toprule[0.8pt]
    \multirow{2}{*}{Method}  & \multicolumn{2}{c}{Origin}  & \multicolumn{2}{c}{\textit{w/.} SS (low)}& \multicolumn{2}{c}{\textit{w/.} SS (high)}& \multicolumn{2}{c}{\textit{w/.} SS (oracle)}\\
    \cmidrule(r){2-3}\cmidrule(r){4-5}\cmidrule(r){6-7}\cmidrule(r){8-9}
    & mIoU & F-score & mIoU & F-score& mIoU & F-score& mIoU & F-score\\
    \midrule
    AVSBench\cite{avsbench}&29.8&35.2&27.3 &29.9&31.5&34.9&36.4&39.0\\
    AVSegformer\cite{avsegformer}&36.7 &42.0&35.4&39.0&39.4&42.5&46.3&48.0\\
  \bottomrule
  \end{tabular}
\end{table}
\subsection{Generalization of Stepping Stones Training Strategy}
\begin{figure}[tb]
  \centering
  \includegraphics[width=1\linewidth]{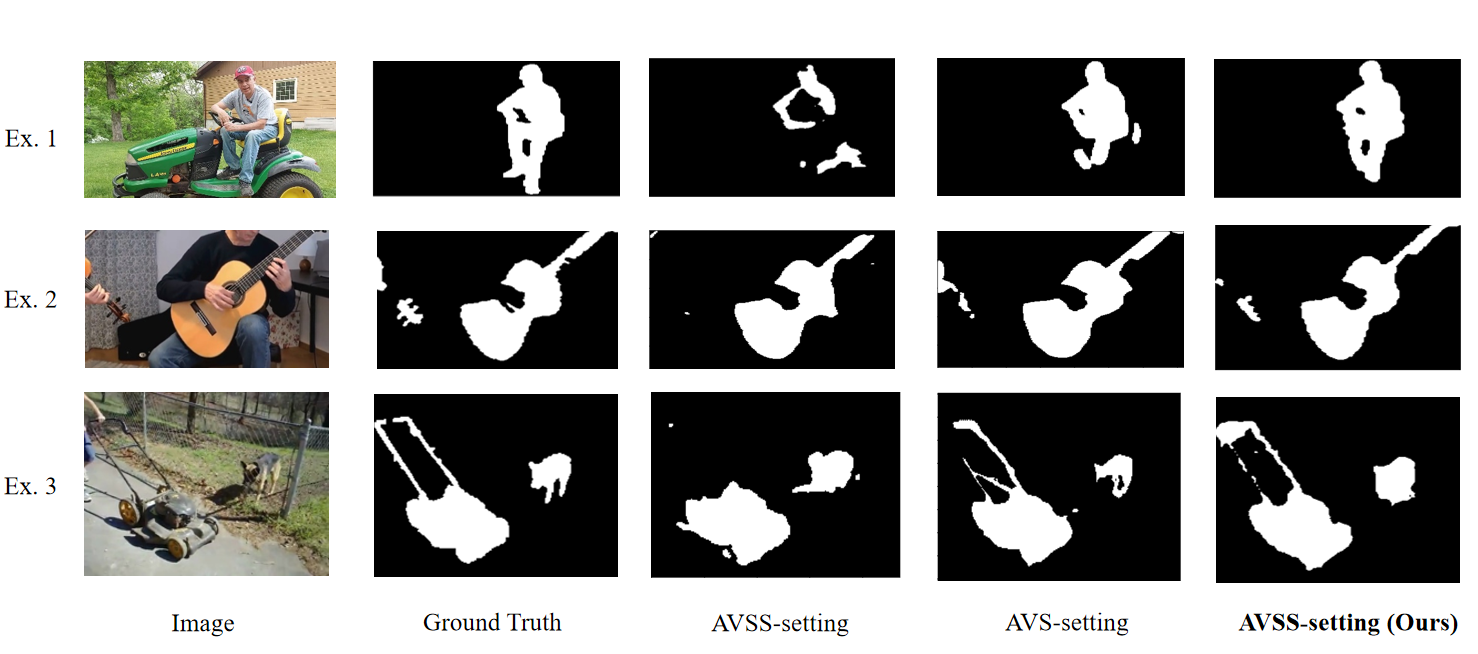}
  \caption{Previous methods exhibit insufficient learning performance under the AVSS setting. The last column represents the model applying the \textit{Stepping Stones} training strategy. Obviously, the model demonstrates inadequate learning when trained in an end-to-end AVSS setting, whereas the localization accuracy of the semantic mask predicted by the model experiences a notable enhancement following the implementation of \textit{Stepping Stones}.}
  \label{fig:compar}
\end{figure}

To assess the effectiveness and generalizability of the \textit{Stepping Stones} training strategy, we also applied it to the AVSBench and AVSegformer models for the AVSS task with minor modifications. Without introducing any special design, we straightforwardly incorporated the sound source localization results with the mask feature in the prediction head, providing a shortcut for the model. Since the effectiveness of the \textit{Stepping Stones} training strategy relies on the accuracy of the first stage results during inferring, we simulated three levels of accuracy of the first stage results as inputs to the second stage in this section. The results are presented in \cref{tab:general}, revealing that when the accuracy of the first stage results is low, the model's performance evens declines due to the dangerous stepping stones. Conversely, when the first stage results demonstrate high accuracy, serving as a reliable stepping stone, the performance of both methods improves. Experiments conducted above only apply the \textit{Stepping Stones} strategy with minimal modifications to the original method. This underscores the possibility of achieving further performance enhancements by exploring alternative approaches to leveraging the first stage results. The enhancement in performance resulting from the application of the \textit{Stepping Stones} strategy to AVSBench is also evident in \cref{fig:compar}.

\begin{table}[htb]
  \caption{Ablation Experiment on adaptive audio queries.}
  \label{tab:ablation1}
  \centering
  \begin{tabularx}{\textwidth}{@{}l *{6}{X}@{}}
        \toprule[0.8pt]
        \multirow{2}{*}{\textbf{Method}}  & \multicolumn{2}{c}{\textbf{S4}}  & \multicolumn{2}{c}{\textbf{MS3}}& \multicolumn{2}{c}{\textbf{AVSS}}\\
        \cmidrule(r){2-3}\cmidrule(r){4-5}\cmidrule(r){6-7}
        & mIoU & F-score & mIoU & F-score& mIoU & F-score\\
    \midrule
    Origin Audio Query  & 80.7&88.4&65.8&75.2&36.0&41.0\\
    Adaptive Audio Query & \textbf{82.5} & \textbf{91.3}&\textbf{67.5}&\textbf{77.6}&\textbf{37.1}&\textbf{41.7}\\
        \bottomrule
    \end{tabularx}
\end{table}

\subsection{Ablation Study}
We conduct ablation experiments to validate the effectiveness of the proposed AAVS model and each key design in the \textit{Stepping Stones} training strategy.

\subsubsection{Ablation of adaptive audio queries.}
We initially assess the effectiveness of the Adaptive Audio Query Generator within the AAVS model. The experiments were conducted across all three subtasks, with the visual backbone frozen. The results are presented in \cref{tab:ablation1}. It is evident that the inclusion of this module enhances the model's performance compared to using the same audio query across all three subtasks.

\subsubsection{Ablation of Stepping Stones training strategy.}
Finally, we conduct experiments to assess the effectiveness of key components in the \textit{Stepping Stones} training strategy. Since the validity of several components in this section is highly correlated with the accuracy of the AVS results provided from the first stage, our experimental results shed light on the impact of actual pseudo labels and ground truth labels. The pseudo labels, inferred from the trained AAVS model, yield an IoU of 83.2\% for S4 labels, 67.5\% for MS3 labels, and 72.8\% for V2 labels. The results of robustly encoding keys and fusing the sound source localization results with mask features are displayed in \cref{tab:ablation4}. It is evident that all components exhibit enhancements compared to the single-stage AAVS model, highlighting the significant improvement brought about by the \textit{Stepping Stones} training strategy.

\begin{table}[tb]
  \caption{Ablation Experiment of \textit{Stepping Stones} training strategy on AAVS. \textit{Actual} refers to sound source localization results inferred from the trained AAVS model, while ``Oracle'' denotes the utilization of ground truth binary labels. Here, the baseline is the AAVS with audio initialization of the attention mask.}
  \label{tab:ablation4}
  \centering
  \begin{tabularx}{\textwidth}{@{}l*{4}{X}@{}}
        \toprule
    \multirow{2}{*}{\textbf{Method}}& \multicolumn{2}{c}{\textbf{Actual}}& \multicolumn{2}{c}{\textbf{Oracle}}\\
    \cmidrule(r){2-3}\cmidrule(r){4-5}
    & mIoU & F-score & mIoU & F-score\\
    \midrule
    final&48.5&53.2&64.9&67.0\\
    \textit{w/o.} mask fusion&45.2 &49.6&54.8&59.7 \\
    \textit{w/o.} robust key&46.4&51.4&60.2&62.3\\
    \textit{w/o.} robust key+mask fusion  &43.4 &48.3&51.2&56.2\\
    \bottomrule
    \end{tabularx}
\end{table}

\section{Conclusion}
In this paper, we propose a simple yet effective progressive training strategy called \textit{Stepping Stones}, where the sound source localization results obtained from the first-stage model serve as a stepping stone for the second-stage model, facilitating a more seamless and comprehensive learning on AVSS subtask. The method is generalizable and directly applicable to preceding end-to-end methods, thereby enhancing their efficacy in AVSS task performance. In the future, the effectiveness of our training strategy is expected to increase as sound source localization methods with greater accuracy emerge. Additionally, we propose a novel framework AAVS designed for dynamically integrating audio and visual features. Extensive experimental results substantiate the superior performance of AAVS and the \textit{Stepping Stones} training strategy compared to existing state-of-the-art methods. For the future work, there is still a large gap between using sound source localization results obtained from the first stage and ground truth labels, as shown in \cref{exper}. The potential of this training strategy remains ripe for exploration, with future endeavors aimed at delving deeper into enhancing the robustness of the model for stepping stones with noise.
\newpage
\section*{Acknowledgements}
This research was supported by National Natural Science Foundation of China (NO.62106272), and Public Computing Cloud, Renmin University of China.

%
%


\begin{thebibliography}{10}
\providecommand{\url}[1]{\texttt{#1}}
\providecommand{\urlprefix}{URL }
\providecommand{\doi}[1]{https://doi.org/#1}

\bibitem{local1}
Arandjelovic, R., Zisserman, A.: Objects that sound. CoRR
  \textbf{abs/1712.06651} (2017), \url{http://arxiv.org/abs/1712.06651}

\bibitem{local5}
Chen, H., Xie, W., Afouras, T., Nagrani, A., Vedaldi, A., Zisserman, A.:
  Localizing visual sounds the hard way. In: CVPR. pp. 16867--16876 (2021)

\bibitem{mask2former}
Cheng, B., Misra, I., Schwing, A.G., Kirillov, A., Girdhar, R.:
  Masked-attention mask transformer for universal image segmentation. In: CVPR.
  pp. 1290--1299 (2022)

\bibitem{maskformer}
Cheng, B., Schwing, A., Kirillov, A.: Per-pixel classification is not all you
  need for semantic segmentation  \textbf{34},  17864--17875 (2021)

\bibitem{dosovitskiy2021imageworth16x16words}
Dosovitskiy, A., Beyer, L., Kolesnikov, A., Weissenborn, D., Zhai, X.,
  Unterthiner, T., Dehghani, M., Minderer, M., Heigold, G., Gelly, S.,
  Uszkoreit, J., Houlsby, N.: An image is worth 16x16 words: Transformers for
  image recognition at scale (2021), \url{https://arxiv.org/abs/2010.11929}

\bibitem{avsegformer}
Gao, S., Chen, Z., Chen, G., Wang, W., Lu, T.: Avsegformer: Audio-visual
  segmentation with transformer (2024)

\bibitem{audioset}
Gemmeke, J.F., Ellis, D.P., Freedman, D., Jansen, A., Lawrence, W., Moore,
  R.C., Plakal, M., Ritter, M.: Audio set: An ontology and human-labeled
  dataset for audio events. In: ICASSP. pp. 776--780. IEEE (2017)

\bibitem{avsbg}
Hao, D., Mao, Y., He, B., Han, X., Dai, Y., Zhong, Y.: Improving audio-visual
  segmentation with bidirectional generation. arXiv preprint arXiv:2308.08288
  (2023)

\bibitem{vggish}
Hershey, S., Chaudhuri, S., Ellis, D.P., Gemmeke, J.F., Jansen, A., Moore,
  R.C., Plakal, M., Platt, D., Saurous, R.A., Seybold, B., et~al.: Cnn
  architectures for large-scale audio classification. In: ICASSP. pp. 131--135.
  IEEE (2017)

\bibitem{adapter}
Houlsby, N., Giurgiu, A., Jastrzebski, S., Morrone, B., De~Laroussilhe, Q.,
  Gesmundo, A., Attariyan, M., Gelly, S.: Parameter-efficient transfer learning
  for nlp. In: ICML. pp. 2790--2799. PMLR (2019)

\bibitem{local2}
Hu, D., Wei, Y., Qian, R., Lin, W., Song, R., Wen, J.R.: Class-aware sounding
  objects localization via audiovisual correspondence  \textbf{44}(12),
  9844--9859 (2021)

\bibitem{aqformer}
Huang, S., Li, H., Wang, Y., Zhu, H., Dai, J., Han, J., Rong, W., Liu, S.:
  Discovering sounding objects by audio queries for audio visual segmentation.
  In: IJCAI. pp. 875--883 (2023)

\bibitem{jain2022oneformertransformerruleuniversal}
Jain, J., Li, J., Chiu, M., Hassani, A., Orlov, N., Shi, H.: Oneformer: One
  transformer to rule universal image segmentation (2022),
  \url{https://arxiv.org/abs/2211.06220}

\bibitem{sam}
Kirillov, A., Mintun, E., Ravi, N., Mao, H., Rolland, C., Gustafson, L., Xiao,
  T., Whitehead, S., Berg, A.C., Lo, W.Y., Dollár, P., Girshick, R.: Segment
  anything (2023)

\bibitem{catr}
Li, K., Yang, Z., Chen, L., Yang, Y., Xiao, J.: Catr: Combinatorial-dependence
  audio-queried transformer for audio-visual video segmentation. In: ACM MM.
  pp. 1485--1494 (2023)

\bibitem{li2024omgsegmodelgoodsegmentation}
Li, X., Yuan, H., Li, W., Ding, H., Wu, S., Zhang, W., Li, Y., Chen, K., Loy,
  C.C.: Omg-seg: Is one model good enough for all segmentation? (2024),
  \url{https://arxiv.org/abs/2401.10229}

\bibitem{fpn}
Lin, T.Y., Doll{\'a}r, P., Girshick, R., He, K., Hariharan, B., Belongie, S.:
  Feature pyramid networks for object detection. In: CVPR. pp. 2117--2125
  (2017)

\bibitem{avsc}
Liu, C., Li, P.P., Qi, X., Zhang, H., Li, L., Wang, D., Yu, X.: Audio-visual
  segmentation by exploring cross-modal mutual semantics. In: ACM MM. pp.
  7590--7598 (2023)

\bibitem{autr}
Liu, J., Ju, C., Ma, C., Wang, Y., Wang, Y., Zhang, Y.: Audio-aware
  query-enhanced transformer for audio-visual segmentation. arXiv preprint
  arXiv:2307.13236  (2023)

\bibitem{SAMA}
Liu, J., Wang, Y., Ju, C., Ma, C., Zhang, Y., Xie, W.: Annotation-free
  audio-visual segmentation. pp. 5604--5614 (2023)

\bibitem{swin}
Liu, Z., Lin, Y., Cao, Y., Hu, H., Wei, Y., Zhang, Z., Lin, S., Guo, B.: Swin
  transformer: Hierarchical vision transformer using shifted windows. In: ICCV.
  pp. 10012--10022 (2021)

\bibitem{fcn}
Long, J., Shelhamer, E., Darrell, T.: Fully convolutional networks for semantic
  segmentation. In: CVPR. pp. 3431--3440 (2015)

\bibitem{diffuse}
Mao, Y., Zhang, J., Xiang, M., Lv, Y., Zhong, Y., Dai, Y.: Contrastive
  conditional latent diffusion for audio-visual segmentation. arXiv preprint
  arXiv:2307.16579  (2023)

\bibitem{vae}
Mao, Y., Zhang, J., Xiang, M., Zhong, Y., Dai, Y.: Multimodal variational
  auto-encoder based audio-visual segmentation. In: CVPR. pp. 954--965 (2023)

\bibitem{dice}
Milletari, F., Navab, N., Ahmadi, S.A.: V-net: Fully convolutional neural
  networks for volumetric medical image segmentation. In: 2016 fourth
  international conference on 3D vision (3DV). pp. 565--571. Ieee (2016)

\bibitem{local3}
Owens, A., Efros, A.A.: Audio-visual scene analysis with self-supervised
  multisensory features. In: ECCV. pp. 631--648 (2018)

\bibitem{local7}
Qian, R., Hu, D., Dinkel, H., Wu, M., Xu, N., Lin, W.: Multiple sound sources
  localization from coarse to fine. In: ECCV. pp. 292--308. Springer (2020)

\bibitem{unet}
Ronneberger, O., Fischer, P., Brox, T.: U-net: Convolutional networks for
  biomedical image segmentation. In: Medical Image Computing and
  Computer-Assisted Intervention--MICCAI 2015: 18th International Conference,
  Munich, Germany, October 5-9, 2015, Proceedings, Part III 18. pp. 234--241.
  Springer (2015)

\bibitem{local4}
Senocak, A., Oh, T.H., Kim, J., Yang, M.H., Kweon, I.S.: Learning to localize
  sound source in visual scenes. In: CVPR. pp. 4358--4366 (2018)

\bibitem{local6}
Song, Z., Wang, Y., Fan, J., Tan, T., Zhang, Z.: Self-supervised predictive
  learning: A negative-free method for sound source localization in visual
  scenes. In: CVPR. pp. 3222--3231 (2022)

\bibitem{gavs}
Wang, Y., Liu, W., Li, G., Ding, J., Hu, D., Li, X.: Prompting segmentation
  with sound is generalizable audio-visual source localizer (2024)

\bibitem{wang2024segpref}
Wang, Y., Sun, P., Li, Y., Zhang, H., Hu, D.: Can textual semantics mitigate
  sounding object segmentation preference? IEEE European Conference on Computer
  Vision (ECCV)  (2024)

\bibitem{wang2024refavs}
Wang, Y., Sun, P., Zhou, D., Li, G., Zhang, H., Hu, D.: Ref-avs: Refer and
  segment objects in audio-visual scenes. IEEE European Conference on Computer
  Vision (ECCV)  (2024)

\bibitem{wei2022learning}
Wei, Y., Hu, D., Tian, Y., Li, X.: Learning in audio-visual context: A review,
  analysis, and new perspective. arXiv preprint arXiv:2208.09579  (2022)

\bibitem{mutr}
Yan, S., Zhang, R., Guo, Z., Chen, W., Zhang, W., Li, H., Qiao, Y., He, Z.,
  Gao, P.: Referred by multi-modality: A unified temporal transformer for video
  object segmentation. AAAI  (2023)

\bibitem{ade}
Zhou, B., Zhao, H., Puig, X., Fidler, S., Barriuso, A., Torralba, A.: Scene
  parsing through ade20k dataset. In: CVPR. pp. 633--641 (2017)

\bibitem{avss}
Zhou, J., Shen, X., Wang, J., Zhang, J., Sun, W., Zhang, J., Birchfield, S.,
  Guo, D., Kong, L., Wang, M., et~al.: Audio-visual segmentation with
  semantics. arXiv preprint arXiv:2301.13190  (2023)

\bibitem{avsbench}
Zhou, J., Wang, J., Zhang, J., Sun, W., Zhang, J., Birchfield, S., Guo, D.,
  Kong, L., Wang, M., Zhong, Y.: Audio--visual segmentation. In: ECCV. pp.
  386--403. Springer (2022)

\end{thebibliography}
\end{document}